\documentclass[runningheads]{llncs}
\usepackage{graphicx}
\usepackage{todonotes}
\usepackage{subcaption}

\usepackage{subcaption}
\usepackage{hyperref}       
\usepackage{url}            
\usepackage{booktabs}       
\usepackage{amsfonts}       
\usepackage{nicefrac}       
\usepackage{microtype}      
\usepackage{siunitx}

\usepackage{multirow}
\begin{document}
\title{Self-distillation for surgical action recognition}
%
%
\author{Amine Yamlahi\inst{1} \and
Thuy Nuong Tran\inst{1} \and
Patrick Godau \inst{1}  \inst{2}\and
Melanie Schellenberg\inst{1} \inst{2}\and
Dominik Michael\inst{1} \inst{2}\and
Finn-Henri Smidt\inst{1}\and
Jan-Hinrich Nölke\inst{1}\and
Tim Adler\inst{1}\and
Minu Dietlinde Tizabi\inst{1}\and
Chinedu Nwoye\inst{3}\and
Nicolas Padoy\inst{3}\and
Lena Maier-Hein\inst{1} \inst{2}\inst{4}\inst{5}}
%

\authorrunning{Amine Yamlahi et al.}
%
\institute{Division of Intelligent Medical Systems, German Cancer Research Center (DKFZ), Heidelberg, Germany \email{m.elyamlahi@dkfz-heidelberg.de} \and
National Center for Tumor Diseases (NCT), Heidelberg, Germany \and
ICube Laboratory, University of Strasbourg, France \and
Faculty of Mathematics and Computer Science, Heidelberg University, Germany \and
Medical Faculty, Heidelberg University, Germany}

%
\maketitle              

\begin{abstract}
Surgical scene understanding is a key prerequisite for context-aware decision support in the operating room. While deep learning-based approaches have already reached or even surpassed human performance in various fields, the task of surgical action recognition remains a major challenge. With this contribution, we are the first to investigate the concept of self-distillation as a means of addressing class imbalance and potential label ambiguity in surgical video analysis. Our proposed method is a heterogeneous ensemble of three models that use Swin Transfomers as backbone and the concepts of self-distillation and multi-task learning as core design choices. According to ablation studies performed with the CholecT45 challenge data via cross-validation, the biggest performance boost is achieved by the usage of soft labels obtained by self-distillation. 
External validation of our method on an independent test set was achieved by providing a Docker container of our inference model to the challenge organizers. According to their analysis, our method outperforms all other solutions submitted to the latest challenge in the field. Our approach thus shows the potential of self-distillation for becoming an important tool in medical image analysis applications.

\keywords{Surgical action recognition \and Self-distillation \and Laparoscopic surgery \and Surgical workflow }
\end{abstract}

\section{Introduction}\label{sec:intro}
\begin{figure}[h]
\centering
\includegraphics[width=0.9\textwidth]{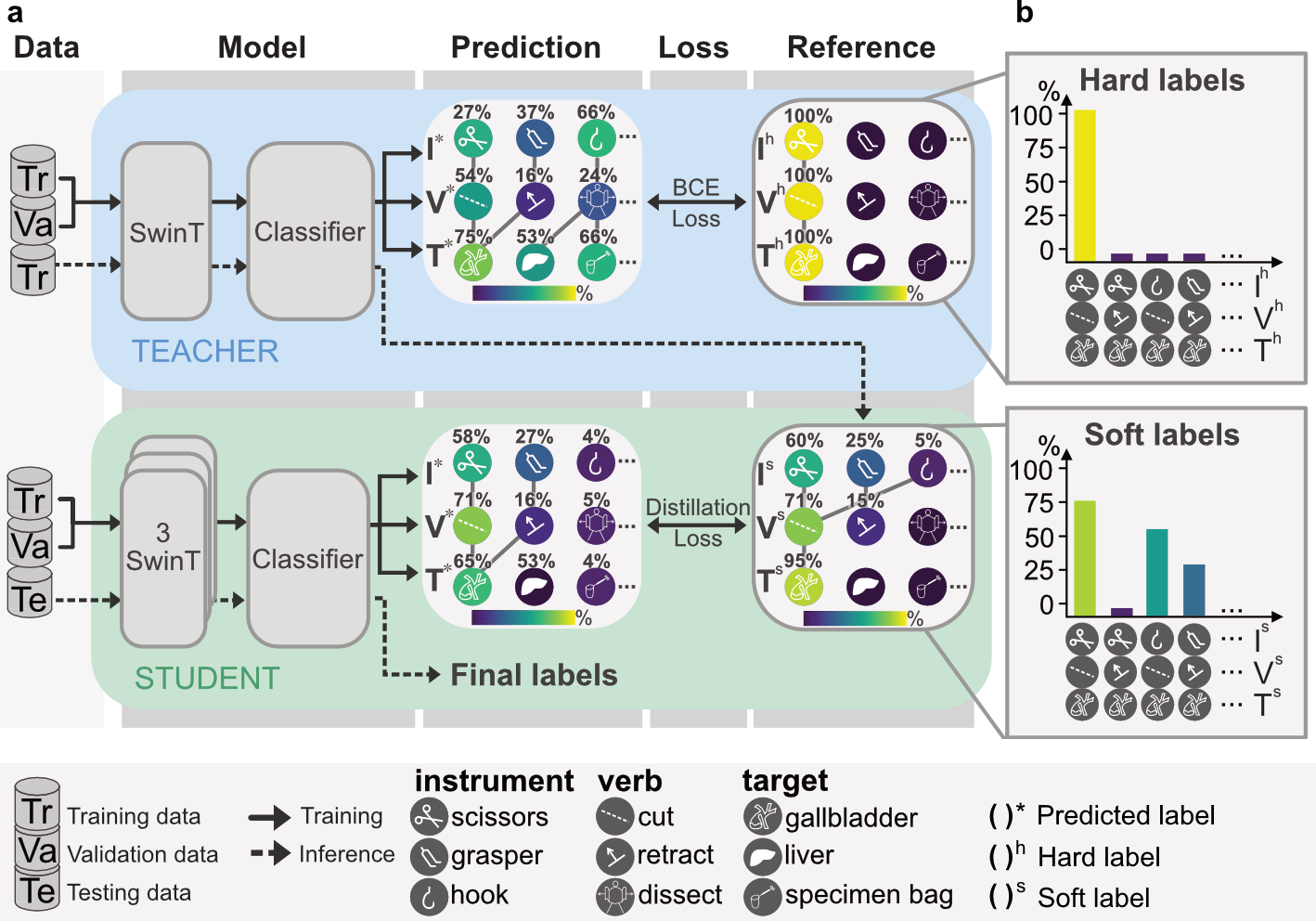}
\caption{\textbf{Approach to surgical action triplet recognition.} (a) Our architecture leverages Swin Transformer (SwinT) as a backbone and the concepts of self-distillation, multi-task learning, and ensembling as core strategies. The teacher model is trained on hard labels using binary cross-entropy (BCE) loss. Inferencing the training data, the sigmoid probabilities are used as input in the next step. The student model is trained with the noisy soft labels in a multi-task fashion to minimize the BCE loss, commonly referred to as distillation loss, between the teacher and the student’s predictions. (b) Visualization of label distribution in hard and soft labels.}
\label{fig:main}
\end{figure}

Surgical scene understanding is an important prerequisite for artificial intelligence (AI)-empowered surgery~\cite{maier2022surgical}, underlying a range of application areas such as context-aware decision support, autonomous robotics, and workflow optimization. One of its key components is the fully-automatic recognition of the surgical action performed at a given point in time – a task not yet solved by state-of-the-art-methods. To advance the field, the CholecTriplet challenge was organized in the scope of the Medical Image Computing and Computer Assisted Interventions (MICCAI) conferences 2021 and 2022. However, according to the organizers analysis \cite{nwoye2022cholectriplet2021,cholectriplet2022web}, the task still remains unsolved. The guiding hypothesis of our work was that self-distillation could address some of the challenges in surgical action recognition, namely the high number of classes (100 in the case of CholecTriplet); high class imbalance, and label ambiguity. Self-distillation builds upon the widespread concept of knowledge distillation (KD)~\cite{hinton2015distilling}, in which the knowledge is transferred from one deep model (i.e., a teacher) to another shallow model (i.e., a student). Self-distillation diverges from traditional KD by distilling knowledge within the network itself. While KD is already used in various communities, the purpose of this work was to pioneer the concept of self-distillation in the context of surgical data science. Based on the CholecTriplet training data set, we developed a method for surgical action recognition (Fig.~\ref{fig:main}) that leverages self-distillation, Swin Transformers~\cite{Liu_2021_ICCV}, multi-task learning, and ensembling as a core design choice. The following Sec.~\ref{sec:methods} presents our methodological contribution in detail. Sec.~\ref{sec:experiments} presents ablation studies on the challenge data set that reveal the most important design choices as well as an external validation of our solution on an independent surgical video data set. We conclude with a brief discussion of the most relevant aspects of our work in Sec.~\ref{sec:discussion}.

\begin{figure}[h]
\centering
\includegraphics[width=1\textwidth]{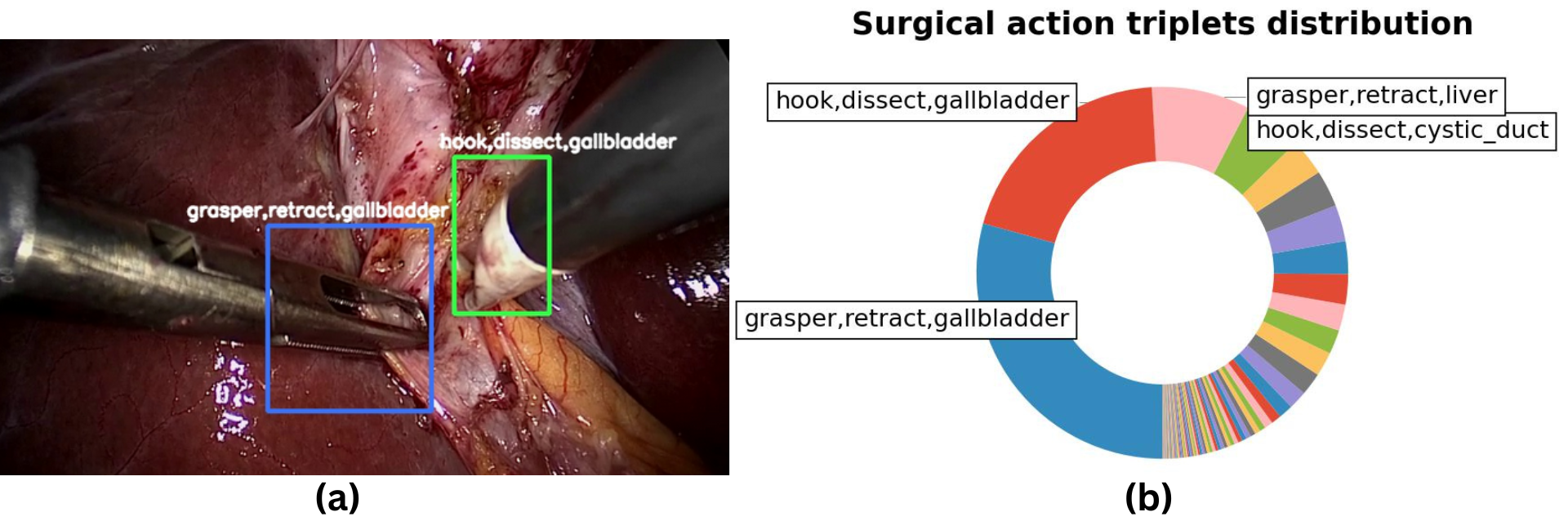}
\caption{\textbf{Task of surgical action recognition.} (a) Each action is represented by a triplet comprising instrument, verb and target. Multiple triplets can be present in one image, as shown in the example. (b) CholecTriplet training data set illustrating the heavy class imbalance. Of 100 possible triplet classes, the prevalence ranges from 0.01\% to 44.6\%.}
\label{fig:overview}
\end{figure}

\section{Methods}\label{sec:methods}

\subsection{Task description and dataset}
Our study is based on the CholecTriplet Challenge 2022~\cite{cholectriplet2022web}, which was conducted under the umbrella of the Endoscopic Vision Challenges (EndoVis) in conjunction with MICCAI. The Surgical Action Recognition task required participants to submit solutions that recognize surgical action triplets in laparoscopic videos, as illustrated in Fig~\ref{fig:overview}. The challenge granted access to the CholecT45~\cite{nwoye2022rendezvous} dataset which consists of 45 video recordings of laparoscopic cholecystectomy with a total of 90,489 frames. CholecT45 is annotated with 100 action triplet classes, with one instrument, verb, and target forming a triplet. The annotations include six different instrument classes, ten verbs (denoting the action performed), and 15 targets such as organs, tissues, or foreign bodies (clip, specimen bag, etc.). The theoretical maximum of \(6\cdot10\cdot15\) classes was reduced to the above-mentioned 100 based on medical relevance and prevalence. An example image from the CholecT45 dataset containing two triplet annotations can be seen in Fig.~\ref{fig:overview} (a). A chart depicting the highly imbalanced class distribution is shown in Fig.~\ref{fig:overview} (b). 

\subsection{Concept overview} 
As illustrated in Fig.~\ref{fig:main}, our approach is based on the following key components:
(1) Swin Transformer: The recently proposed Swin Transformer~\cite{Liu_2021_ICCV} architecture was chosen as backbone. 
(2) Multi-task learning: Based on the success of previous work that leveraged multi-task learning as its training paradigm, we incorporated multiple auxiliary tasks in our architecture, namely the classification of the individual components of the triplet (instrument, verb, and target) as well as the surgical phase.
(3) Self-distillation: The core idea of our approach is the usage of soft labels to reduce overconfidence and address label ambiguity.
(4) Ensemble: Following common successful training strategies, we implement ensembling to combine the predictions of three trained Swin Transformers of different scales.
\\

\subsection{Implementation details}
\textbf{Swin Transformer} We base our method on Swin Transformer (SwinT) models of the timm~\cite{rw2019timm} library and adopted the final classification layer to output the 100 triplet predictions, as well as the individual instruments, verbs, targets and surgical phase as auxiliary tasks to leverage the interconnection between them in a multi-task fashion (+Multi).\\
\textbf{Self-Distillation} The concept of self-distillation was achieved by training a teacher Swin transformer on one-hot encoded hard labels for 20 epochs, with a batch size of 64, an  Adam~\cite{kingma2014adam} optimizer, a learning rate of \num{2e-4}, a cosine annealing scheduler decreasing to a minimum learning rate of \num{2e-6}, and a binary cross-entropy loss function. The model was trained with light augmentations that comprise resizing the images to 224x224 pixels, horizontal and vertical flips, rotation, brightness and saturation perturbations with a probability of 0.5. We trained five teacher and five student models; one for each fold of the official five-fold cross validation splitting introduced by the challenge. The teacher was trained on four of the five splits of its fold. After convergence, the soft labels (i.e., the sigmoid probabilites) for the same four splits were computed and the student was trained using these soft labels. The validation was performed on the fifth split, using hard (i.e., the original) labels for both the teacher and the student. During inference, the sigmoid probabilities of the five student models were averaged to yield the final result. 
The five teacher models shared a common weight intialization seed. The five student models shared a separate weight initialization seed. The student models were trained for 40 epochs with the same augmentations as the teacher models. We saved the weights on the epoch with the best mean Average Precision (mAP) score based on the validation split for the current fold.\\
\textbf{Ensemble} We combined three trained Swin Transformers (SwinT) of different scales (SwinT base/SwinT large) and configurations for our final ensemble (Ens) model: First, we employed a SwinT base model with multi-task learning of instrument, verb, and target and trained it using self-distillation. Second, we used a SwinT large model using the same approach, and added label smoothing to the soft labels. Third, we included phase annotations as an additional task for the multi-task training of a SwinT base model still employing self-distillation. Please note that every single model mentioned here corresponds to the five aggregated student models of the previous paragraph.

\section{Experiments and results}\label{sec:experiments}
The purpose of the experiments was to validate the performance of our method and to quantify the (potential) benefit of each individual component. To this end, we conducted (1) comprehensive ablation studies using the CholecT45 official 5-Fold cross-validation split~\cite{nwoye2022data}, (2) an analysis of the specific benefit of soft labels, and (3) an external validation based on a Docker container submitted to the CholecTriplet 2022 challenge organizers. The Rendezvous Net~\cite{cholectriplet2022web}, provided by the challenge organizers, served as the benchmark. In line with the challenge design~\cite{miccaisig}, we validated the performance using mAP (following the aggregation scheme in \cite{nwoye2022cholectriplet2021}) and the top K=5 Accuracy as metrics. All scores were computed using the ivtmetrics library~\cite{nwoye2022data}.

\subsubsection{Ablation studies} 
\begin{figure}[ht]
\centering
\includegraphics[width=0.9\textwidth]{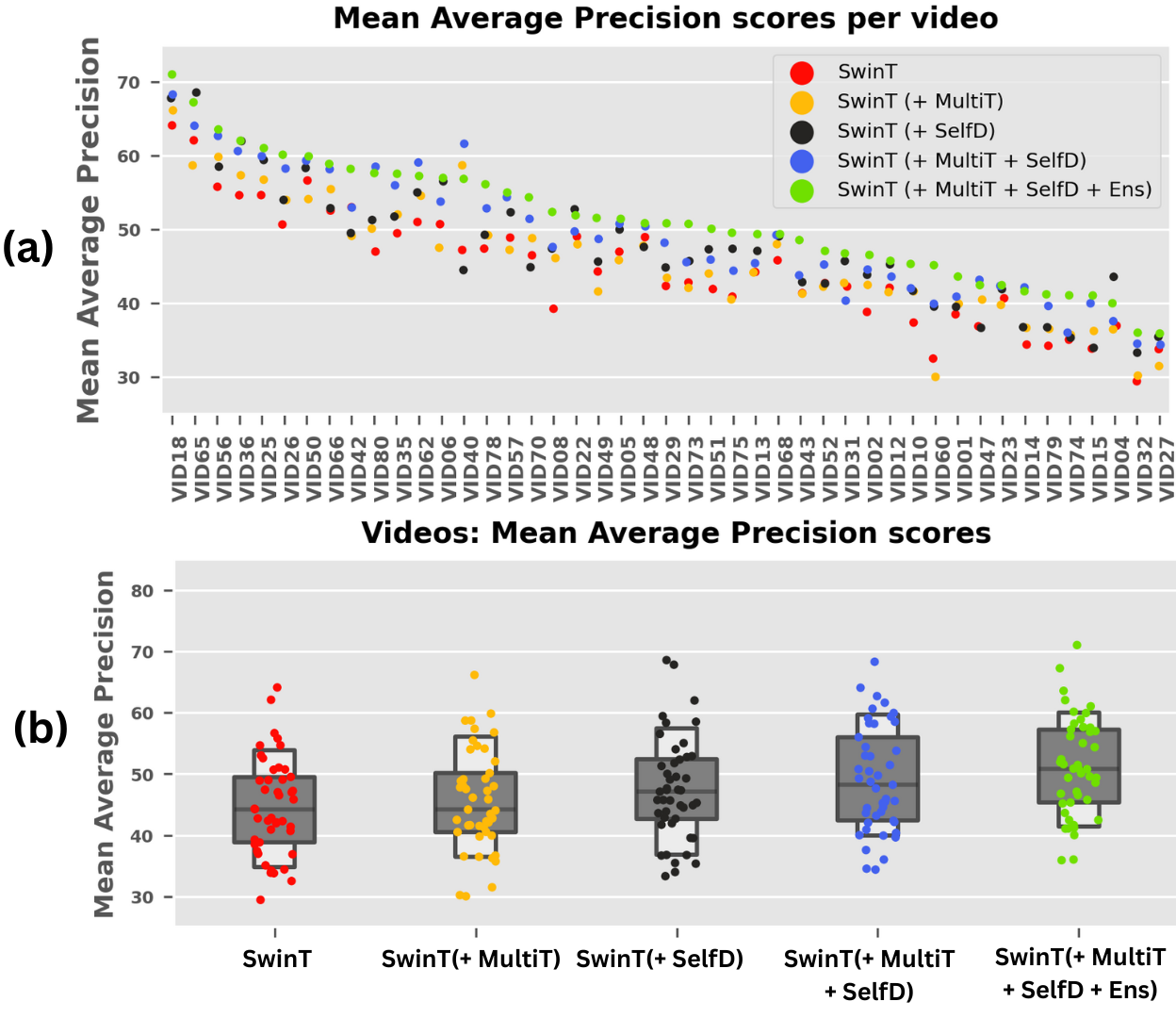}
\caption{\textbf{Quantitative results on the validation data.} (a) mean Average Precision (mAP) plotted separately for each video for five configurations of our method with increasing complexity. A Swin Transformer (SwinT) was gradually complemented by  multi-task learning (MultiT), self-distillation (SelfD), and ensembling (Ens). Videos were sorted by mAP score of final ensemble performance from highest to lowest. (b) Corresponding dot- and boxplots of mAP scores.  
}\label{fig:videos}
\end{figure}

We designed the ablation studies as follows: We first calculated the performance of our Swin Transformer backbone as a stand-alone triplet classifier (SwinT). Next, we added multiple auxiliary targets (instruments, verbs, targets, and phases) for multi-task classification (+MultiT). As a third component, we implemented self-distillation by training a student model on soft labels, acquired by training the teacher model (+SelfD). The fourth step was the ensembling of three student model Swin Transformers (+Ens). The results are shown in Tab.~\ref{tab:ablation_seg}. A single SwinT as model backbone yields a higher mAP for triplet classification (mAP=32.3\%) than the benchmark (mAP=28.8\%), which corresponds to a relative improvement by 10.3\%. The biggest boost was achieved by including self-distillation, which improved the Triplet mAP and top-5 accuracy by 3.8 percentage points (pp) and 2.4pp, respectively, compared to our baseline. The final model yielded a mAP of 38.5\% and a top-5 accuracy of 86.5\%, which corresponds to a boost of 6.2pp in mAP and 2.7pp in top-5 accuracy compared to our own baseline, and a relative improvement by 33.7\% for mAP compared to the state-of-the-art method.  For transparency, we also provide per-video results, depicted in Fig.~\ref{fig:videos}. With a few exceptions, our final model consistently provides the best results.

\begin{table}[]
\resizebox{\textwidth}{!}{%
\begin{tabular}{l|cc|cc|}
\cline{2-5}
\multirow{2}{*}{}                                                                                     & \multicolumn{2}{c|}{Cross Validation}                                                                                                                    & \multicolumn{2}{c|}{\begin{tabular}[c]{@{}c@{}}External validation\\ (CholecTriplet 2022 test set)\end{tabular}}                                         \\ \cline{2-5} 
                                                                                                      & \multicolumn{1}{c|}{\begin{tabular}[c]{@{}c@{}}Triplet\\ mAP  {[}\%{]}\end{tabular}} & \begin{tabular}[c]{@{}c@{}}Top-5\\ accuracy {[}\%{]}\end{tabular} & \multicolumn{1}{c|}{\begin{tabular}[c]{@{}c@{}}Triplet\\ mAP  {[}\%{]}\end{tabular}} & \begin{tabular}[c]{@{}c@{}}Top-5\\ accuracy {[}\%{]}\end{tabular} \\ \hline
\multicolumn{1}{|l|}{Rendezvous}                                                                      & \multicolumn{1}{c|}{28.8}                                                            & N/A                                                               & \multicolumn{1}{c|}{32.7}                                                            & 69.3                                                              \\ \hline
\multicolumn{1}{|l|}{Ours: SwinT}                                                                     & \multicolumn{1}{c|}{32.3}                                                            & 83.8                                                              & \multicolumn{1}{c|}{32.9}                                                            & 70.7                                                              \\ \hline
\multicolumn{1}{|l|}{Ours: SwinT + MultiT}                                                            & \multicolumn{1}{c|}{33.1}                                                            & 84.6                                                              & \multicolumn{1}{c|}{33.8}                                                            & 71.6                                                              \\ \hline
\multicolumn{1}{|l|}{Ours: SwinT + SelfD}                                                             & \multicolumn{1}{c|}{35.0}                                                            & 85.2                                                              & \multicolumn{1}{c|}{36.1}                                                            & 72.9                                                              \\ \hline
\multicolumn{1}{|l|}{\begin{tabular}[c]{@{}l@{}}Ours: SwinT + MultiT\\ + SelfD\end{tabular}}          & \multicolumn{1}{c|}{36.1}                                                            & 86.2                                                              & \multicolumn{1}{c|}{37.3}                                                            & 73.3                                                              \\ \hline
\multicolumn{1}{|l|}{\begin{tabular}[c]{@{}l@{}}Ours: SwinT + MultiT\\ \\ + SelfD + Ens\end{tabular}} & \multicolumn{1}{c|}{\textbf{38.5}}                                                   & \textbf{86.5}                                                     & \multicolumn{1}{c|}{\textbf{37.4}}                                                   & \textbf{74.0}                                                     \\ \hline
\end{tabular}%
}
\caption{\textbf{Main quantitative results} Starting from our backbone model – a
Swin Transformer (Swin T) – we gradually added individual components, namely
multi-task learning (MultiT), self-distillation (SelfD), and ensembling (Ens). Each
component addition leads to an increase in mAP and top-5 accuracy, in both
cross-validation (left) and independent external validation (right). N/A: Not
available.}
\label{tab:ablation_seg}
\end{table}

\subsubsection{Analysis of soft labels}
\begin{figure}[h]
\centering
\includegraphics[width=0.8\textwidth]{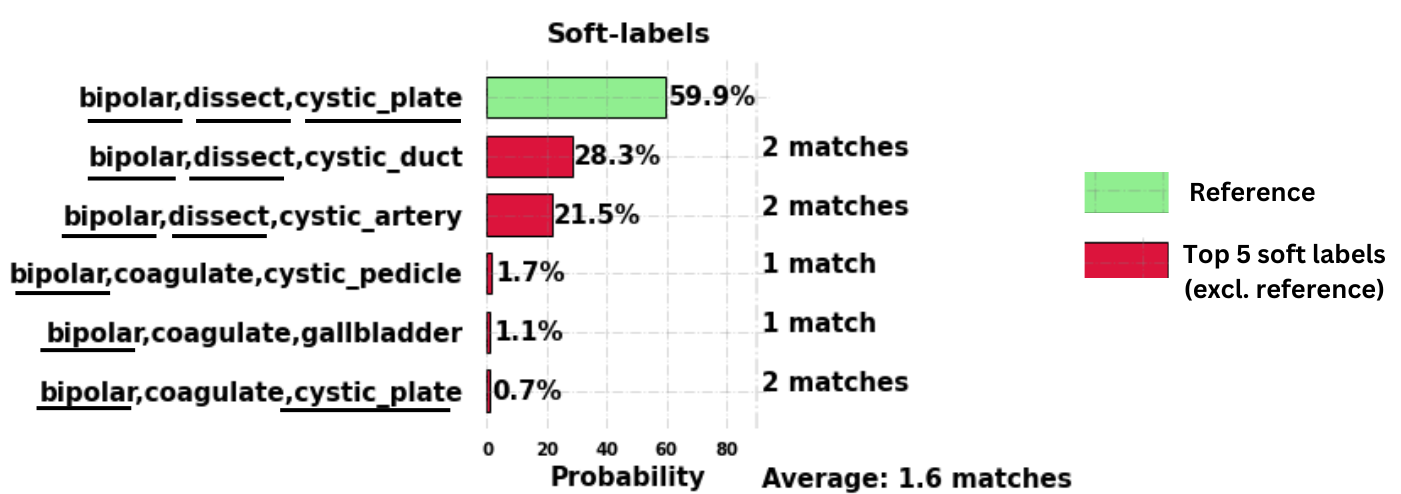}
\caption{\textbf{Example of generated soft labels} with reference (in green) and top 5 soft label triplets ranked by probability (in red). Average number of component matches between reference “bipolar, dissect, cystic\_plate” and top five triplets is 1.6. 
}\label{fig:softlabel}
\end{figure}

\begin{figure}[h]
\centering
\includegraphics[width=0.8\textwidth]{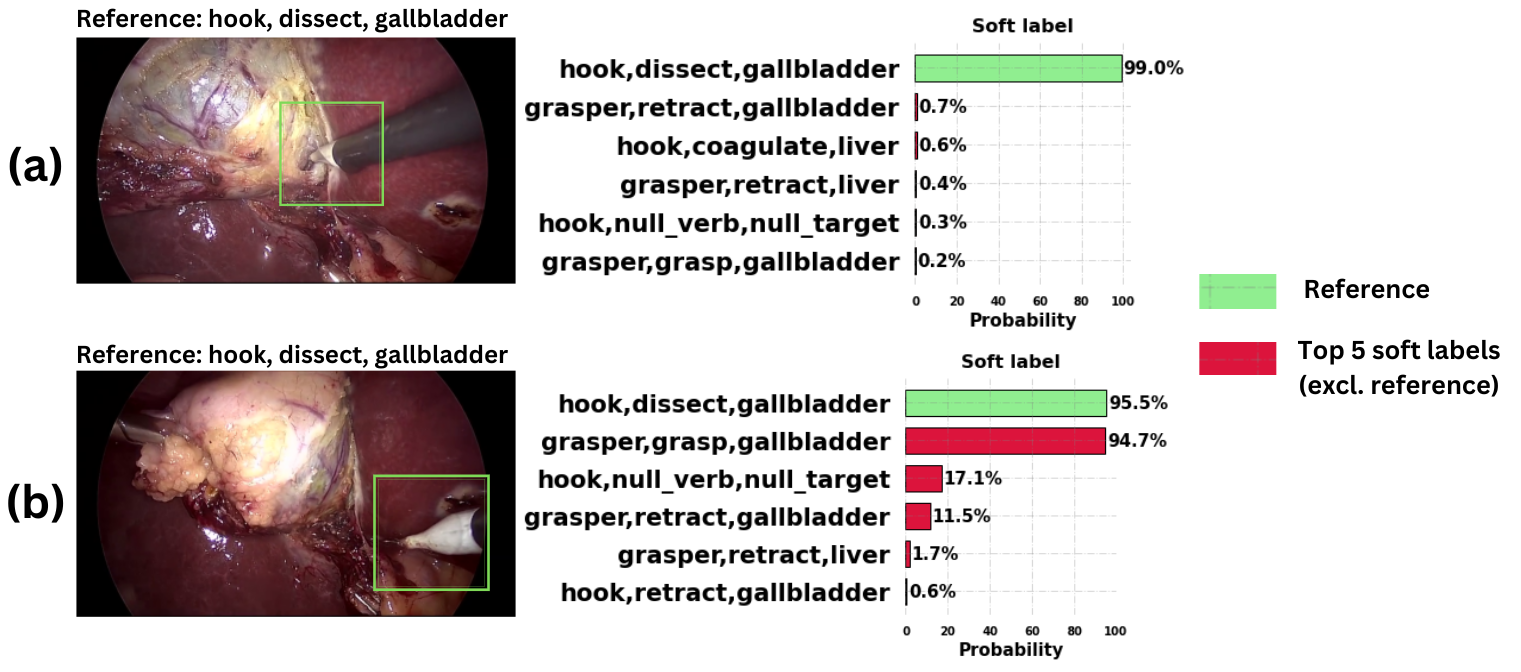}
\caption{\textbf{Examples of training images with corresponding soft labels}. (a) Single triplet with corresponding unambiguous (confident) soft label (b) Case with a poor reference label. The missing triplet  “grasper, grasp, gallbladder” is compensated for in the soft label annotation, leading to a more complete reference in the actual network training achieved by self-distillation.}
\label{fig:suppl1}
\end{figure}

The addition of self-distillation resulted in the highest boost in performance. This holds true despite the fact that the mAP of the teacher model, trained on hard labels, was about 88\% on the training set, which is suboptimal. The question is thus why the poorer soft labels still yielded a performance improvement. While part of the answer is provided in the literature on soft/noisy labels~\cite{kim2021self,mobahi2020self,vu2021teaching,yun2020regularizing}, we also speculated that the soft labels may address the issue of ambiguous/erroneous labels in our particular use case. More specifically, we assumed that in the case of a faulty (wrong) annotation, the reference and ground truth label might be semantically close to each other. Hence, if the student model increases the scores of semantically close triplets, it might actually increase the value for the ground truth label, which might explain the performance boost. To investigate whether this is the case, we first defined a pragmatic proxy metric for semantic similarity: the number of identical triplet items (max: two for different triplets). We then selected all frames with only one unique triplet label and retrieved the top five triplets (excluding the reference) with the highest soft label score. Fig.~\ref{fig:softlabel}, depicts an example of such a comparison. The reference triplet “bipolar, dissect, cystic\_plate” is shown with five soft label triplets ranked by probability. In the example, the top five triplets share an average of 1.6 components with the reference, indicating that they contain similar semantic information. 
We found that over all samples, the average number of component matches between reference and top five triplets is 1.0$\pm0.002$. In contrast, when comparing the reference with five triplets randomly drawn (while respecting prevalence), the agreement is 0.5$\pm0.002$.
This shows that self-distillation specifically leads to increased scores for semantically related classes. 
Fig ~\ref{fig:suppl1} shows examples of training images with corresponding soft labels.

\subsubsection{Independent external validation}
External validation was conducted on the CholecTriplet challenge test set. The results, shown in Tab.~\ref{tab:ablation_seg}, confirm the results from cross-validation experiments, with the final ensemble scoring 37.4\% in mAP. This equals an absolute improvement of 4.7pp and a relative improvement of 14.4\% compared to the Rendezvous benchmark (mAP= 32.7\%). According to the challenge organizers, our method outperformed all other solutions submitted in 2022 by at least 2.9pp (Triplet mAP) and 8.3pp (top-5 accuracy) absolute.

\section{Discussion}\label{sec:discussion}
This paper pioneers the concept of self-distillation in the medical image analysis domain. Specifically, we are the first to tackle key challenges in surgical action recognition, namely the high number of classes and class imbalance, with self-distillation. Comprehensive ablation studies combined with external validation yielded the following findings: 
\begin{enumerate}
    \item  Swin Transformers, as recently introduced by the computer vision community, can serve as a strong backbone in endoscopic vision tasks. This is suggested by the fact that even our most ablated model, consisting of a single Swin Transformer, surpasses the state-of-the-art surgical action recognition method \textit{Rendezvouz}.
    \item Multi-task learning, here using the classification of instrument, verb, and target as well as of the surgical phase as auxiliary tasks, yielded a notable increase in performance.
    \item Self-distillation yielded the biggest boost in performance, suggesting that soft labels are better suited for surgical action recognition.  

 \item Ensembling increased performance further, as also suggested by various publications in a wide range of fields. 
\end{enumerate}


Overall, the addition of self-distillation (in combination with the Swin Transformer as a backbone) resulted in the highest performance boost. While label noise has been shown to be beneficial in various work~\cite{kim2021self,mobahi2020self,vu2021teaching,yun2020regularizing}, the concept of self-distillation may not necessarily be intuitive; although the mAP achieved by the teacher model, trained on hard labels, is suboptimal (32\%), the teacher's noisy labels lead to an overall improvement in performance when compared to the (presumably better-quality) hard labels. In the general machine learning literature, the knowledge encoded in noisy labels is referred to as "dark knowledge" because it is not yet well-understood. Aiming to shed light on this topic, our experiments on semantic similarity suggest that soft labels may actually address the issue of ambiguous/erroneous labels. Further analyses with more sophisticated metrics for semantic similarity are, however, needed to support this finding.

Related work has so far tackled the challenge of surgical action recognition with various strategies including multi-task learning~\cite{nwoye2020recognition,ramesh2021multi}, and different attention mechanisms~\cite{czempiel2021opera,nwoye2022rendezvous} incorporated into diverse architectures based on temporal convolutional networks~\cite{czempiel2020tecno,ramesh2021multi}, transformers~\cite{czempiel2021opera,gao2021trans,nwoye2022rendezvous}, or combinations of convolutional neural networks (CNN) with recurrent neural networks (RNN)~\cite{jin2017sv,jin2021temporal,nwoye2020recognition,yu2018learning} or hidden Markov models (HMM)~\cite{twinanda2016endonet}. While our approach was particularly successful according to the challenge analysis, the overall performance is still not optimal. Advancing the methods will require more data that features a sufficient number of samples for each triplet and captures the full variability of scenes that might be encountered in practice. From a methodological perspective, future work should be directed to efficiently taking temporal context into account and addressing potential domain shifts~\cite{castro2020causality}.

In conclusion, our study is the first to demonstrate the benefit of self-distillation for surgical vision tasks. Based on the substantial performance boost obtained, the usage of soft labels could become a valuable tool in the endoscopic vision community.

\section{Acknowledgments}\label{sec:acknowledgments}
This project was supported by a Twinning Grant of the German Cancer Research Center (DKFZ) and the Robert Bosch Center for Tumor Diseases (RBCT). Part of this work was  funded by the Surgical Oncology Program of the National Center for Tumor Diseases (NCT) Heidelberg and by the German Federal Ministry of Health under the reference number 2520DAT0P1 as part of the pAItient project, and by HELMHOLTZ IMAGING, a platform of the Helmholtz Information \& Data Science Incubator and by French state funds managed within the Plan Investissements d’Avenir by the ANR under references: National AI Chair AI4ORSafety [ANR-20-CHIA-0029-01], Labex CAMI [ANR-11-LABX-0004], DeepSurg [ANR-16-CE33-0009], IHU Strasbourg [ANR-10-IAHU-02] and by BPI France under references: project CONDOR, project 5G-OR. 
Model Docker evaluation were performed with servers/HPC resources managed by CAMMA, IHU Strasbourg, Unistra Mésocentre, and GENCI-IDRIS [Grant 2021-AD011011638R1, 2021-AD011011638R2, 2021-AD011011638R3].

\newpage
\bibliographystyle{splncs04}
\bibliography{main}

\end{document}